\documentclass[runningheads]{llncs}
\usepackage[T1]{fontenc}
\usepackage{times}
\usepackage{epsfig}
\usepackage{graphicx}
\usepackage{amsmath}
\usepackage{amssymb}

\usepackage{wrapfig}

\usepackage{booktabs}

\usepackage{url}
\usepackage{subcaption}
\usepackage{multirow}
\usepackage{comment}
\usepackage{balance}
\usepackage{pifont}
\usepackage{arydshln}
\usepackage{algorithm, algpseudocode}
\usepackage{amsfonts}

\usepackage{adjustbox}
\usepackage{xcolor,colortbl}
\usepackage[normalem]{ulem}

\DeclareMathOperator*{\argmin}{arg\,min}

\definecolor{Gray}{gray}{0.5}

\definecolor{LightCyan}{rgb}{0.88,1,0.88}

\newcommand{\algstep}[1]{\item[]\medskip\hrule\kern 2pt\hbox to \textwidth{\hspace{\labelsep}\textbf{#1}\hfill}\hrule}

\usepackage[pagebackref=true,breaklinks=true,letterpaper=true,colorlinks,bookmarks=false]{hyperref}

\graphicspath{ {imgs/} }
\usepackage{dsfont}
\usepackage{mathtools}
\usepackage{booktabs}
\usepackage{graphics}
\usepackage[mathscr]{euscript}

\renewcommand{\vec}[1]{{\mathbf #1}}
\newcommand{\mat}[1]{{\mathbf #1}}

\makeatletter
\DeclareRobustCommand\onedot{\futurelet\@let@token\@onedot}
\def\@onedot{\ifx\@let@token.\else.\null\fi\xspace}
\def\eg{\emph{e.g}\onedot} 

\def\ie{\emph{i.e}\onedot}

\def\etal{\emph{et al}\onedot}
\makeatother

\renewcommand{\vec}[1]{{\mathbf #1}}
\usepackage{pifont}

\usepackage{hyperref}
\hypersetup{
    colorlinks=true,
    linkcolor=blue,
    filecolor=magenta,      
    urlcolor=cyan,
}
\usepackage[font=scriptsize]{caption}
\usepackage{mwe}
\usepackage[capitalize]{cleveref}
\crefname{section}{Sec.}{Secs.}
\Crefname{section}{Section}{Sections}
\Crefname{table}{Table}{Tables}

\begin{document}
\title{L3DMC: Lifelong Learning using Distillation via Mixed-Curvature Space}
\titlerunning{L3DMC}

\author{Kaushik Roy %
\inst{1, 2}, Peyman Moghadam %
\inst{2}, Mehrtash Harandi %
\inst{1}}
\authorrunning{K. Roy et al.}
\institute{Department of Electrical and Computer Systems Engineering, Faculty of Engineering, Monash University, Melbourne, Australia. \and  Data61, CSIRO, Brisbane, QLD, Australia.\\
\email{\{Kaushik.Roy, Mehrtash.Harandi\}@monash.edu \\
\{Kaushik.Roy, Peyman.Moghadam\}@csiro.au}}

\maketitle       

\renewcommand*{\thefootnote}{\fnsymbol{footnote}}
\footnotetext[0]{Manuscript has been accepted in MICCAI2023 (Early Accept)}
\vspace{-5mm}
\begin{abstract}
The performance of a lifelong learning (L3) model degrades when it is trained on a series of tasks, as the geometrical formation of the embedding space changes while learning novel concepts sequentially. The majority of existing L3 approaches operate on a fixed-curvature (\eg, zero-curvature Euclidean) space that is not necessarily suitable for modeling the complex geometric structure of data. Furthermore, the distillation strategies apply constraints directly on low-dimensional embeddings, discouraging the L3 model from learning new concepts by making the model highly stable. To address the problem, we propose a distillation strategy named L3DMC that operates on mixed-curvature spaces to preserve the already-learned knowledge by modeling and maintaining complex geometrical structures. We propose to embed the projected low dimensional embedding of fixed-curvature spaces (Euclidean and hyperbolic) to higher-dimensional Reproducing Kernel Hilbert Space (RKHS) using a positive-definite kernel function to attain rich representation. Afterward, we optimize the L3 model by minimizing the discrepancies between the new sample representation and the subspace constructed using the old representation in RKHS. L3DMC is capable of adapting new knowledge better without forgetting old knowledge as it combines the representation power of multiple fixed-curvature spaces and is performed on higher-dimensional RKHS. %
Thorough experiments on three benchmarks demonstrate the effectiveness of our proposed distillation strategy for medical image classification in L3 settings. Our code implementation is publicly available at \href{https://github.com/csiro-robotics/L3DMC}{https://github.com/csiro-robotics/L3DMC}.

\end{abstract}

\vspace{-4.50ex}
\section{Introduction}
\label{sec:intro}
\vspace{-2.5mm}
Lifelong learning~\cite{parisi2019continual,ring1997child} is the process of sequential learning  from a series of non-stationary data distributions through acquiring novel concepts while preserving already-learned knowledge. However, \underline{D}eep \underline{N}eural \underline{N}etworks (DNNs) exhibit a significant drop in performance on previously seen tasks when trained in continual learning settings. This phenomenon is often called catastrophic forgetting~\cite{mccloskey1989catastrophic,nguyen2019toward,robins1995catastrophic}. Furthermore, the unavailability of sufficient training data in medical imaging poses an additional challenge in tackling catastrophic forgetting of a DNN model.

In a lifelong learning scenario, maintaining a robust embedding space and preserving geometrical structure is crucial to mitigate performance degradation and catastrophic forgetting of old tasks\cite{knights2022incloud}. However, the absence of samples from prior tasks has been identified as one of the chief reasons for catastrophic forgetting.  Therefore, to address the problem, a small memory buffer has been used in the literature to store a subset of samples from already seen tasks and replayed together with new samples~\cite{rebuffi2017icarl,hou2019learning}. Nevertheless,  imbalances in data (\eg, between current tasks and the classes stored in the memory) make the model biased towards the current task~\cite{hou2019learning}.
\begin{wrapfigure}{t}{6.75cm}
\scriptsize
\vspace{-5mm}
\hspace{-5mm}
\centering
    \includegraphics[height=4.25cm, width=1.05\linewidth]{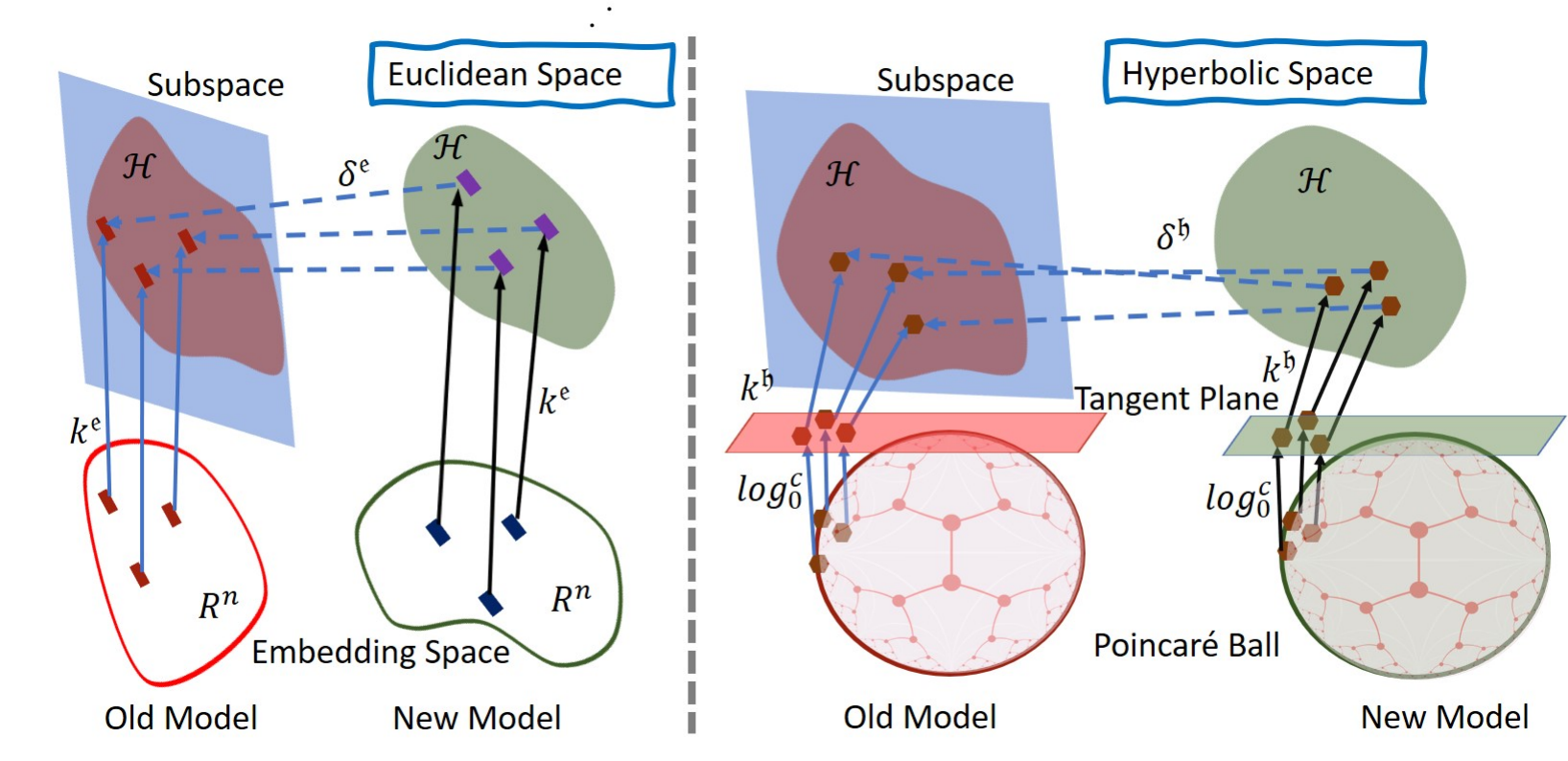}
    \caption{ Geometrical interpretation of the proposed distillation strategy, L3DMC via mixed-curvature space that optimizes the model by combining the distillation loss from Euclidean with zero-curvature (left) and hyperbolic with negative-curvature (right) space. L3DMC preserves the complex geometrical structure by minimizing the distance between new data representation and subspace induced by old representation in RKHS.}
    \label{fig:l3dmc}
    \vspace{-7mm}
\end{wrapfigure}
Knowledge distillation~\cite{hinton2015distilling,parisi2019continual,douillard2020podnet,buzzega2020dark} has been widely used in the literature to preserve the previous knowledge while training on a novel data distribution. This approach applies constraints on updating the weight of the current model by mimicking the prediction of the old model. To maintain the old embedding structure intact in the new model,  feature distillation strategies have been introduced in the literature~\cite{hou2019learning,douillard2020podnet}. For instance, LUCIR~\cite{hou2019learning} emphasizes on maximizing the similarity between the orientation of old and new embedding by minimizing the cosine distance between old and new embedding. %
While effective, feature distillation applies strong constraints directly on the lower-dimensional embedding extracted from the old and new models, reducing the plasticity of the model. This is not ideal for adopting novel concepts while preserving old knowledge. 
Lower-dimensional embedding spaces, typically used for distillation, may not preserve all the latent information in the input data~\cite{jayasumana2015kernel}. As a result, they may not be ideal for distillation in lifelong learning scenarios. Furthermore, DNNs often operate on zero-curvature (\ie, Euclidean) spaces, which may not be suitable for modeling and distilling complex geometrical structures
in non-stationary biomedical image distributions with various modalities and discrepancies in imaging protocols and medical equipment. On the contrary, hyperbolic spaces have been successfully used to model hierarchical structure in input data for different vision tasks~\cite{khrulkov2020hyperbolic,ganea2018hyperbolic}.

In this paper,  we propose to perform distillation in a Reproducing Kernel Hilbert Space (RKHS), constructed from the embedding space of multiple fixed-curvature spacess. This approach is inspired by the ability of kernel methods to yield rich representations in 
higher-dimensional RKHS~\cite{fang2021kernel,jayasumana2015kernel}. Specifically, we employ a \underline{R}adial \underline{B}asis \underline{F}unction (RBF) kernel on a mixed-curvature space that combines embeddings from hyperbolic (negative curvature), and Euclidean (zero curvature), using a decomposable Riemannian distance function as illustrated in Fig.~\ref{fig:l3dmc}. This mixed-curvature space is robust and can maintain a higher quality geometrical formation. This makes the space more suitable for knowledge distillation and tackling catastrophic forgetting in lifelong learning scenarios for medical image classification. Finally, to ensure a similar geometric structure between the old and new models in L3, we propose minimizing the distance between the new embedding and the subspace constructed using the old embedding in RKHS.
Overall, our contributions in this paper are as follows:
\begin{itemize}
    \item To the best of our knowledge, this is the first attempt to study mixed-curvature space for the continual medical image classification task.
    \item We propose a novel knowledge distillation strategy to maintain a similar geometric structure for continual learning by minimizing the distance between new embedding and subspace constructed using old embedding in RKHS.
    \item Quantitative analysis shows that our proposed distillation strategy is capable of preserving complex geometrical structure in embedding space resulting in significantly less degradation of the performance of continual learning  and superior performance compared to state-of-the-art baseline methods on BloodMNIST, PathMNIST, and OrganaMNIST datasets.
\end{itemize}

\vspace{-2.0ex}
\section{Preliminaries}
\label{sec:preliminaries}

\textbf{Lifelong Learning (L3).} L3 consists of a series of $T$ tasks $\mathcal{T}_t \in \{\mathcal{T}_1, \mathcal{T}_2, \cdots, \mathcal{T}_T\} $, where each task $\mathcal{T}_t$ has it's own dataset $\mathcal{D}_t = \{\mathcal{X}_t, \mathcal{Y}_t\}$. 
In our experiments, $\mat{X}_i \in \mathcal{X}_t \subset \mathcal{X}$ denotes a medical image of size $W \times H$ and  $y_i \in \mathcal{Y}_t \subset \mathcal{Y}$ is its associated disease category at task $t$. In class-incremental L3, label space of two tasks is disjoint, hence $\mathcal{Y}_{t} \cap \mathcal{Y}_{t'} = \emptyset; t \neq t'$. The aim of L3 is to  train a model $f: \mathcal{X} \to \mathcal{Y}$ incrementally for each task $t$ to map the input space $\mathcal{X}_t$ to the corresponding target space $\mathcal{Y}_t$ without forgetting all previously learned tasks (\ie, $1, 2, \cdots, t-1$). We assume that a fixed-size memory $\mathcal{M}$ is available to store a subset of previously seen samples to mitigate catastrophic forgetting in L3. 
\\
\noindent
\textbf{Mixed-Curvature Space.} Mixed-Curvature space is formulated as the Cartesian product of fixed-curvature spaces and represented as $M = \bigtimes_{i=1}^C M_i^{d_i}$. Here, $M_i$ can be a Euclidean (zero curvature), hyperbolic (constant negative curvature), or spherical (constant positive curvature) space. %
Furthermore, $\bigtimes$ denotes the Cartesian product, and $d_i$ is the dimensionality of fixed-curvature space $M_i$ with curvature $c_i$. The distance in the mixed-curvature space can be decomposed as $d_M(\vec{x}, \vec{y}) \coloneqq \sum_{i=1}^{C} d_{M_i}(\vec{x}^i, \vec{y}^i)$.

\noindent
\textbf{Hyperbolic Poincar\'e Ball.} Hyperbolic space %
is a Riemannian manifold with negative curvature. The Poincare ball with curvature $-c,~ c>0$, $\mathbb{D}_c^n = \{\vec{x} \in \mathbb{R}^n : c\|\vec{x}\| < 1 \}$ is a model of $n$-dimensional hyperbolic geometry. To perform vector operations on $\mathbb{H}^n$, M$\ddot{\text{o}}$bius Gyrovector space is widely used. M$\ddot{\text{o}}$bius addition between $x \in \mathbb{D}_c^n$ and $y \in \mathbb{D}_c^n$ is defined as follows
\begin{align}
    \vec{x} \oplus_c \vec{y} = \frac{
            (1 + 2 c \langle \vec{x}, \vec{y}\rangle + c \|\vec{y}\|^2_2) \vec{x} + (1 - c \|\vec{x}\|_2^2) \vec{y}
            }{
            1 + 2 c \langle \vec{x}, \vec{y}\rangle + c^2 \|\vec{x}\|^2_2 \|\vec{y}\|^2_2
        }
\end{align}
Using M$\ddot{\text{o}}$bius addition, geodesic distance between two input data points, $\vec{x}$ and $\vec{y}$ in $\mathbb{D}_c^n$ is computed using the following formula.
\begin{align}
\label{eq:geodesic_h}
    d_c(\vec{x}, \vec{y}) = \frac{2}{\sqrt{c}}\tanh^{-1}(\sqrt{c}\|(-\vec{x})\oplus_c \vec{y}\|_2)
\end{align}
Tangent space of data point $\vec{x} \in \mathbb{D}_c^n$ is the inner product space and is defined as $T_x\mathbb{D}_c^n$ which comprises the tangent vector of all directions at $\vec{x}$. Mapping hyperbolic embedding to Euclidean space and vice-versa is crucial for performing operations on $\mathbb{D}^n$. Consequently, a vector $\vec{x} \in T_x\mathbb{D}_c^n$ is embedded onto the Poincar\'e ball $\mathbb{D}^n_c$ with anchor $\vec{x}$ using the exponential mapping function and 
and the inverse process is done using the logarithmic mapping function $\mathrm{log}_{\vec{v}}^c$ that maps $\vec{x} \in \mathbb{D}_c^n$ to the tangent space of $\vec{v}$ as follows 
\begin{align}
\label{eq:log_map}
    \log_{\vec{v}}^c(\vec{x}) = \frac{2}{\sqrt{c}\lambda_{\vec{v}}^c} \tanh^{-1}( 
            \sqrt{c} \|-\vec{v} \oplus_c \vec{x}\|_2)  \frac{-\vec{v} \oplus_c \vec{x}}{\|-\vec{v}\oplus_c \vec{x}\|_2}
\end{align}
where $\lambda_{\vec{v}}^c$ is conformal factor that is defined as $\lambda_{\vec{v}}^c=\frac{2}{1-c\|v\|^2}$. In practice, anchor $\vec{v}$ is set to the origin. Therefore, the exponential mapping is expressed as $\mathrm{exp}_0^c(\vec{x}) = \mathrm{tanh}\big(\sqrt{c}\|\vec{x}\|\big)\frac{\vec{x}}{\sqrt{c}\|\vec{x}\|}$.

\vspace{-2.0ex}
\section{Proposed Method}
\label{sec:methodology}
\vspace{-2.0ex}
In our approach, we emphasize on modeling complex
latent structure of medical data by combining embedding representation of zero-curvature Euclidean and negative-curvature hyperbolic space. To attain richer representational power of RKHS~\cite{hofmann2008kernel}, we embed the low-dimensional fixed-curvature embedding onto higher-dimensional RKHS using the kernel method. %

\noindent
\textbf{Definition 1.} \textit{\big(Positive Definite Kernel\big)} A function $f: \mathcal{X}\times\mathcal{X} \rightarrow \mathbb{R}$ is \underline{p}ositive \underline{d}efinite (pd) if and only if \textbf{1.} $k(\vec{x_i}, \vec{x_j}) = k(\vec{x_j}, \vec{x_i})$ for any $\vec{x_i}, \vec{x_j} \in \mathcal{X}$, and \textbf{2.} for any given $n \in \mathbb{N}$, we have $\sum_i^n\sum_j^n c_ic_j k(\vec{x_j}, \vec{x_i}) \geq 0$ for any $\vec{x}_1, \vec{x}_2, \cdots, \vec{x}_n \in \mathcal{X}$ and  $c_1, c_2, \cdots, c_n \in \mathbb{R}$. Equivalently, the Gram matrix $K_{ij} = k(\vec{x}_i,\vec{x}_j) > 0$ for any set of $n$ samples $\vec{x}_1, \vec{x}_2, \cdots , \vec{x}_n \in \mathcal{X}$ should be Symmetric and Positive Definite (SPD). 

\noindent
Popular kernel functions (\eg, the Gaussian RBF) operate on flat-curvature Euclidean spaces. 
In $\mathbb{R}^n$, the Gaussian RBF kernel method is defined as
\begin{align}
\label{eq:rbf_e}
    k^{\mathfrak{e}}(\vec{z_i}, \vec{z_j}) \coloneqq \mathrm{exp}\big( -\lambda \| \vec{z_i} - \vec{z_j} \|^2 \big); \lambda > 0.
\end{align}
However, using the geodesic distance in a hyperbolic space along with an RBF function similar to \cref{eq:rbf_e} (\ie, replacing 
$\| \vec{z_i} - \vec{z_j} \|^2$ with the geodesic distance) does not lead to a valid positive definite kernel. 
Theoretically, a valid RBF kernel is impossible to obtain for hyperbolic space using geodesic distance~\cite{feragen2015geodesic,feragen2016open}. Therefore,  we use the tangent plane of hyperbolic space and employ $\text{log}_0^c$
\begin{align}
    \mathrm{log}_0^c(\vec{z}) = \frac{\vec{z}}{\sqrt{c}\|\vec{z}\|}\mathrm{tanh}^{-1}\big( \sqrt{c}\|\vec{z}\| \big)\;,
\end{align}
to embed hyperbolic data to RKHS via the following valid pd kernel (see~\cite{fang2021kernel} for the proof of positive definiteness):
\begin{align}
    k^{\mathfrak{h}}(\vec{z_i}, \vec{z_j}) = \mathrm{exp}\big( -\lambda\| \mathrm{log}_0^c(\vec{z_i}) - \mathrm{log}_0^c(\vec{z_j}) \|^2 \big).
\end{align}
\noindent
Now, in L3 setting, we have two models $h^t$ and $h^{t-1}$ at our hand at time $t$. 
We aim to improve $h^t$ while ensuring the past knowledge incorporated in $h^{t-1}$ is kept within $h^t$.
Assume $\vec{Z}_{t}$ and $\vec{Z}_{t-1}$ are the extracted feature vectors for input $\vec{X}$ using current and old feature extractor, $h_{feat}^t$ and $h_{feat}^{t-1}$, respectively. Unlike other existing distillation methods, we employ an independent 2-layer MLP for each fixed-curvature space to project extracted features to a new lower-dimensional embedding space on which we perform further operations. This has two benefits, \textbf{(i)} it relaxes the strong constraint directly applied on $\vec{Z}_{t}$ and $\vec{Z}_{t-1}$ and  \textbf{(ii)} reduce the computation cost of performing kernel method. Since we are interested in modeling embedding structure in zero-curvature Euclidean and negative-curvature hyperbolic spaces, we have two MLP as projection modules attached to feature extractors, namely $g_{e}$ and $g_{h}$.

\paragraph{\textbf{Our idea.}} Our main idea is that, for a rich and overparameterized representation, the data manifold is low-dimensional.
Our algorithm makes use of RKHS, which can be intuitively thought of as a neural network with infinite width. Hence, we assume that the data manifold for the model at time $t-1$ is well-approximated by a low-dimensional hyperplane (our data manifold assumption). Let 
$\vec{Z}_{t-1}^{\mathfrak{e}} = \{\vec{z}^{\mathfrak{e}}_{t-1,1}, \vec{z}^{\mathfrak{e}}_{t-1,2}, \cdots, \vec{z}^{\mathfrak{e}}_{t-1,m}\}$
be the output of the Euclidean projection module for $m$ samples at time $t$ (\ie, current model). 
Consider $\vec{z}^{\mathfrak{e}}_{t}$,  a sample at time $t$  from the Euclidean projection head. We propose to minimize the following distance
\begin{align}
    \label{eqn:kernel_dist_subspace1}
    \delta^{\mathfrak{e}}(\vec{z}^{\mathfrak{e}}_{t}, \vec{Z}_{t-1}^{\mathfrak{e}}) & \coloneqq 
    \Big\| \phi(\vec{z}^{\mathfrak{e}}_{t}) - \text{span}\{ \phi(\vec{z}^{\mathfrak{e}}_{t-1,i}) \}_{i=1}^m \Big \|^2 
    \hspace{-2mm} \\ \notag & = 
    \min_{\alpha \in \mathbb{R}^m} \Big\| \phi(\vec{z}^{\mathfrak{e}}_{t}) - \sum_{i=1}^m \alpha_i \phi(\vec{z}^{\mathfrak{e}}_{t-1,i})\Big \|^2\;.
\end{align}

In~\cref{eqn:kernel_dist_subspace1}, $\phi$ is the implicit mapping to the RKHS defined by the Gaussian RBF kernel, \ie $k^{\mathfrak{e}}$. The benefit of formulation~\cref{eqn:kernel_dist_subspace1} is that it has a closed-form solution as
\begin{align}
    \label{eqn:kernel_dist_subspace2}
    \delta^{\mathfrak{e}}(\vec{z}^{\mathfrak{e}}_{t}, \vec{Z}_{t-1}^{\mathfrak{e}}) = 
    k(\vec{z}^{\mathfrak{e}}_{t},\vec{z}^{\mathfrak{e}}_{t}) - k_{\vec{z}\vec{Z}}^\top{K^{-1}_{\vec{Z}\vec{Z}}}k_{\vec{z}\vec{Z}}\;.
\end{align}
In~\cref{eqn:kernel_dist_subspace2},  $K_{\vec{Z}\vec{Z}} \in \mathbb{R}^{m \times m}$ is the Gram matrix of 
$\vec{Z}_{t-1}^{\mathfrak{e}}$, and   $k_{\vec{z}\vec{Z}}$ is an m-dimensional vector storing the kernel values between
$\vec{z}^{\mathfrak{e}}_{t}$ and elements of $\vec{z}^{\mathfrak{e}}_{t-1}$. We provide the proof of equivalency between 
\cref{eqn:kernel_dist_subspace1} and \cref{eqn:kernel_dist_subspace2} in the supplementary material due to the lack of space. 
Note that we could use the same form for the hyperbolic projection module $g_{h}$ to distill between the model at time $t$ and $t-1$, albeit this time, we employ the hyperbolic kernel $k^{\mathfrak{h}}$. Putting everything together, 
\begin{align}
    \label{eqn:total_kernel_distill_loss}
    \ell_{\text{KD}}(\vec{Z}_t) \coloneqq  \mathbb{E}_{\vec{z}_t} 
    \delta^{\mathfrak{e}}(\vec{z}^{\mathfrak{e}}_{t}, \vec{Z}_{t-1}^{\mathfrak{e}}) + 
    \beta \mathbb{E}_{\vec{z}_t}\delta^{\mathfrak{h}}(\vec{z}^{\mathfrak{h}}_{t}, \vec{Z}_{t-1}^{\mathfrak{h}}) \;. 
\end{align}
Here, $\beta$ is a hyper-parameter that controls the weight of distillation between the Euclidean and hyperbolic spaces.
We can employ~\cref{eqn:total_kernel_distill_loss} at the batch level. Note that in our formulation, computing the inverse of an $m\times m$
matrix is required, which has  a complexity of $\mathcal{O}(m^3)$. However, this needs to be done once per batch and manifold 
(\ie, Euclidean plus hyperbolic). $\ell_{\text{KD}}$ is differentiable with respect to $\vec{Z}_t$, which enables us to update the model at time $t$. We train our lifelong learning model by combining distillation loss, $\ell_{\text{KD}}$, together with standard cross entropy loss.
Please refer to the overall steps of training lifelong learning model using our proposed distillation strategy via mixed-curvature space in Algorithm~\ref{alg:CL3DMC}.

\vspace{-2.0ex}
\subsection{Classifier and Exemplar Selection}
We employ herding based exemplar selection method that selects examples that are closest to the class prototype, following iCARL~\cite{rebuffi2017icarl}. At inference time, we use exemplars from memory to compute class template and the nearest template class computed using Euclidean distance is used as the prediction of our L3 model. Assume $\vec{\mu}_{c}$ is the class template computed by averaging the extracted features from memory exemplars belonging to class $c$. Then, the prediction $\hat{y}$ for a given input sample $\vec{X}$ is determined as $\hat{y}=\underset{c=1,\ldots,t}{\argmin} \|h_{feat}^t(\vec{X}) - \mu_c\|_2$.

\vspace{-1.0ex}
\section{Related Work}
\label{sec:relatedwork}

\begin{wrapfigure}{R}{0.55\textwidth}
\vspace{-16ex}
\begin{minipage}{0.55\textwidth}
\begin{algorithm}[H] 
\caption{Lifelong Learning using Distillation via Mixed-Curvature Space}
\label{alg:CL3DMC}
\begin{algorithmic}[1]
\Require{$\text{Dataset}~\mathcal{D}^0,\mathcal{D}^1,...,\mathcal{D}^T, \text{ and Memory}~\mathcal{M}$}
\Ensure{The new model at time $t$ with parameters $\vec{\Theta}^t$}
\State {Randomly Initialize $\vec{\Theta}^0$;~$h^0_{\vec{\Theta}} = h^0_{\text{feat}} \circ h^0_{\text{cls}}$}
\State {Train $\vec{\Theta}^0$ on $\mathcal{D}^0$ using $\ell_{CE}$}
    \For {$t$  in  $\{1, 2, ... , T\}$}
    \State {$\text{Initialize }\vec{\Theta}^t~\text{ with }\vec{\Theta}^{t-1}$}
    \For {iteration $1$  to  max\_iter}

        \State {Sample a mini batch $(\mathcal{X}_B, \mathcal{Y}_B)$ from $({\mathcal{D}^t} \cup {\mathcal{M}})$}
                
        \State $\vec{Z}^t \leftarrow h_{\text{feat}}^t(\mathcal{X}_B)$
        \State $\vec{Z}^{t-1} \leftarrow h_{\text{feat}}^{t-1}(\mathcal{X}_B)$
        
        \State {$ \mathcal{\tilde{Y}}_B \leftarrow h_{\Theta^t}(\vec{Z}^t)$}
        
        \State {Compute $\ell_{\mathrm{CE}}$ between $\mathcal{Y}_B$ and  $\mathcal{\tilde{Y}}_B$}
        \State {Compute $\ell_{\text{KD}}$ between $\vec{Z}^{t-1}$ and $\vec{Z}^{t}$ %
        }

        \State {Update $\vec{\Theta}^t$ by minimizing the combined loss of cross-entropy $\ell_{\mathrm{CE}}$ and $\ell_{\mathrm{KD}}$ as in Eq.~\eqref{eqn:total_kernel_distill_loss}
        }
    \EndFor
    \State {Evaluate Model on test dataset}
    \State {Update Memory $\mathcal{M}$ with exemplars from $\mathcal{D}^t$}
    \EndFor
\end{algorithmic}
\end{algorithm}
\end{minipage}
\vspace{-3ex}
\end{wrapfigure}

In this section, we describe Mixed-curvature space and L3 methods to tackle catastrophic forgetting.

\noindent
\textbf{Constant-Curvature and Mixed-Curvature Space.} Constant-curvature spaces have been successfully used in the literature to realize the intrinsic geometrical orientation of data for various downstream tasks in machine learning. Flat-curvature Euclidean space is suitable to model grid data~\cite{wu2020comprehensive} while positive and negative-curvature space is better suited for capturing cyclical~\cite{bachmann2020constant} and hierarchical~\cite{liu2019hyperbolic} structure respectively. Hyperbolic representation has been used across domains ranging from image classification~\cite{mathieu2019continuous} and natural language processing~\cite{nickel2017poincare,nickel2018learning} to graphs~\cite{chami2019hyperbolic}. However, a constant-curvature space is limited in modeling the geometrical structure of data embedding as it is designed with a focus on particular structures~\cite{gu2019learning}.

\noindent
\textbf{Kernel Methods.} A Kernel is a function that measures the similarity between two input samples. The intuition behind the kernel method is to embed the low-dimensional input data into a higher, possibly infinite, dimensional RKHS space. Because of the ability to realize rich representation in RKHS, kernel methods have been studied extensively in machine learning~\cite{hofmann2008kernel}. 

\noindent
\textbf{L3 using Regularization with Distillation.}
Regularization-based approaches impose constraints on updating weights of L3 model to maintain the performance on old tasks. LwF mimics the prediction of the old model into the current model but struggles to maintain consistent performance in the absence of a task identifier. Rebuff \etal in~\cite{rebuffi2017icarl} store a subset of exemplars using a herding-based sampling strategy and apply knowledge distillation on output space like LwF~\cite{li2017learning}. Distillation strategy on feature spaces has also been studied in the literature of L3. Hou \etal in~\cite{hou2019learning} proposes a less-forgetting constraint that controls the update of weight by minimizing the cosine angle between old and new embedding representation.

\section{Experimental Details}
\label{sec:experiments}

\noindent
\textbf{Datasets.} In our experiments, we use four datasets (\eg, BloodMNIST~\cite{acevedo2020dataset}, PathMNIST~\cite{kather2019predicting}, OrganaMNIST~\cite{bilic2023liver}) and TissueMNIST~\cite{bilic2023liver} from MedMNIST collection~\cite{yang2021medmnist} for the multi-class disease classification. BloodMNIST, PathMNIST, OrganaMNIST and TissueMNIST have 8, 9, 11, and 8 distinct classes, respectively that are split into 4 tasks with non-overlapping classes between tasks following~\cite{derakhshani2022lifelonger}. For cross-domain continual learning experiments, we present 4 datasets sequentially to the model. %

\noindent
\textbf{Implementation Details.} We employ ResNet18~\cite{he2016deep} as the backbone for feature extraction and a set of task-specific fully connected layers as the classifier to train all the baseline methods across datasets. To ensure fairness in comparisons, we run each experiment with the same set of hyperparameters as used in~\cite{derakhshani2022lifelonger} for five times with a fixed set of distinct seed values, ${1, 2, 3, 4, 5}$ and report the average value. Each model is optimized using Stochastic Gradient Decent (SGD) with a batch of 32 images for 200 epochs, having early stopping options in case of overfitting. Furthermore, we use gradient clipping by enforcing the maximum gradient value to 10 to tackle the gradient exploding problem.

\noindent
\textbf{Evaluation Metrics.} We rely on average accuracy and average forgetting to quantitatively examine the performances of lifelong learning methods as used in previous approaches~\cite{rebuffi2017icarl,chaudhry2018riemannian}. 
Average accuracy is computed by averaging the accuracy of all the previously observed and current tasks after learning a current task $t$ and defined as:
$\text{Acc}_{t} = \frac{1}{t}\sum _{i=1}^{t} \text{Acc}_{t, i},$
where $\text{Acc}_{t, i}$ is the accuracy of task $i$ after learning task $t$.
We measure the forgetting of the previous task at the end of learning the current task $t$ using:
$F_{t} = \frac{1}{t-1}\sum _{i=1}^{t-1} \max_{j \in \{1...t-1\}} \text{Acc}_{j,i} - \text{Acc}_{t, i},$
where at task $t$, forgetting on task $i$ is defined as the maximum difference value previously achieved accuracy and current accuracy on task $i$.

\vspace{-2ex}
\section{Results and Discussion}
\label{sec:results}

In our comparison, we consider two regularization-based methods (\ie, EWC~\cite{kirkpatrick2017overcoming}, and LwF~\cite{li2017learning}) and 5 memory-based methods (\eg, EEIL~\cite{castro2018end}, ER~\cite{riemer2018learning}, Bic~\cite{wu2019large}, LUCIR~\cite{hou2019learning} and iCARL~\cite{rebuffi2017icarl}). We employ the publicly available code\footnote{https://github.com/mmderakhshani/LifeLonger} of~\cite{derakhshani2022lifelonger} in our experiments to produce results for all baseline methods on BloodMNIST, PathMNIST, and OrganaMNIST datasets and report the quantitative results in Tab.~\ref{tab:benchmark}. The results suggest that the performance of all methods improves with the increase in buffer size (\eg, from 200 to 1000). We observe that our proposed distillation approach outperforms other baseline methods across the settings. The results suggest that the regularization-based methods, \eg, EWC and LwF perform poorly in task-agnostic settings across the datasets as those methods are designed for task-aware class-incremental learning. Our proposed method outperforms experience replay, ER method by a significant margin in both evaluation metrics (\ie, average accuracy and average forgetting) across datasets. For instance, our method shows around 30\%, 30\%, and 20\% improvement in accuracy compared to ER while the second best method, iCARL, performs about 4\%, 2\%, and 8\% worse than our method on BloodMNIST, PathMNIST, and OrganaMNIST respectively with 200 exemplars. Similarly, with 1000 exemplars, our proposed method shows consistent performances and outperforms iCARL by 4\%, 6\%, and 2\% accordingly on BloodMNIST, PathMNIST, and OrganaMNIST datasets. We also observe that catastrophic forgetting decreases with the increase of exemplars. Our method shows about 2\% less forgetting phenomenon across the datasets with 1000 exemplars compared to the second best method iCARL.

\begin{table*}[tp!]
\vspace{-.25em}
\centering
\caption{\label{tab:benchmark} Experimental results on BloodMNIST, PathMNIST, OrganaMNIST and TissueMNIST datasets for 4-tasks Class-Incremental setting  with varying buffer size. Our proposed method outperforms other baseline methods across the settings.
}
\begin{adjustbox}{max width=\textwidth}
\begin{tabular}{ccccccc}
\hline
\multirow{1}{*}{Method}                           & \multicolumn{2}{c}{BloodMNIST} 
& \multicolumn{2}{c}{PathMNIST} 
& \multicolumn{2}{c}{OrganaMNIST} 
\\ \cline{2-7}
&     Accuracy~$\uparrow$ & Forgetting~$\downarrow$ & Accuracy~$\uparrow$      & Forgetting~$\downarrow$ 
&     Accuracy~$\uparrow$ & Forgetting~$\downarrow$ \\ \hline
Upper Bound   &  97.98  &  -    &   93.52   &   -   &  95.22   &   -   \\
Lower Bound   &  46.59  &  68.26    &   32.29   &   77.54   &  41.21   &   54.20   \\ \hline
EWC~\cite{kirkpatrick2017overcoming}   &  47.60  &  66.22    &  33.34   &  76.39   &  37.88   &  67.62   \\
LwF~\cite{li2017learning}   &  43.68  &  66.30    &  35.36   &  67.37   &  41.36   &  51.47   \\ 
\hline
\multicolumn{7}{c}{Memory Size: 200} \\  \cline{2-7}
EEIL~\cite{castro2018end}  & 42.17   &  71.25    &  28.42   &  79.39   &   41.03  &  62.47   \\ 
ER~\cite{riemer2018learning}    & 42.94   &  71.38    &  33.74   &  80.6    &   52.50  &  52.72   \\ 
LUCIR~\cite{hou2019learning} &  20.76  &  53.80    &  40.00   &  54.72   &  41.70   &  33.06   \\ 
BiC~\cite{wu2019large}   & 53.32   &  31.06    &  48.74   &  30.82   &  58.68   & 29.66    \\ 
iCARL~\cite{rebuffi2017icarl} &  67.70  &  \textbf{14.52}    &  58.46   &  \textbf{-0.70}   &  63.02   &  \textbf{7.75}    \\ 
\rowcolor{LightCyan}
Ours   &  \textbf{71.98}   &  14.62   &  \textbf{60.60}   &  21.18   &  \textbf{71.01}   &  13.88  \\ 
\hline
\multicolumn{7}{c}{Memory Size: 1000} \\  \cline{2-7}
EEIL~\cite{castro2018end}  &  64.40  &   40.92   &   34.18  &  75.42   &  66.24   &  34.60   \\ 
ER~\cite{riemer2018learning}    &  65.94  &   33.68   &  44.18   &  66.24   &  67.90   &  31.72   \\ 
LUCIR~\cite{hou2019learning} &  20.92  &   28.42   &  53.84   &  30.92   &   54.22  &  23.64   \\ 
BiC~\cite{wu2019large}   &  70.04  &  17.98    &  -   &   -  &  73.46   &  15.98   \\ 
iCARL~\cite{rebuffi2017icarl} &  73.10  &  13.18    &  61.72   &   14.14  &  74.54   &  10.50   \\ 
\rowcolor{LightCyan}
Ours  &  \textbf{77.26}   &  \textbf{10.9}   &  \textbf{67.52}   &  \textbf{12.5}   &  \textbf{76.46}   &  \textbf{9.08}   \\ 
\hline

\end{tabular}
\end{adjustbox}
\end{table*}

\begin{table}[t!]
\vspace{-1.75ex}    
    \centering
    \caption{\label{tab:cross_domain} Average accuracy on the cross-domain incremental learning scenario~\cite{derakhshani2022lifelonger} with 200 exemplars. CL3DMC outperforms baseline methods by a significant margin in both task-aware and task-agnostic settings. Best values are in bold.}
    \begin{tabular}{|c|c|c|c|c|c|c|c|c|}
    \hline
Scenario & LwF & EWC & ER & EEIL & BiC & iCARL & LUCIR & L3DMC(Ours) \\    \hline
    Task-Agnostic (Accuracy~$\uparrow$)  &  29.45  & 18.44  &  34.54 &  34.54  & 26.79  &  48.87  & 19.05  &  \textbf{52.19}   \\ \hline
    Task-Aware  (Accuracy~$\uparrow$)     &  31.07  &  29.26 &  37.69 &  33.19 &  33.19  &  49.47 &  27.48 &  \textbf{52.83}   \\ 
    \hline
    
    \end{tabular}
\vspace{-5.0ex}    
\end{table}
Tab.~\ref{tab:cross_domain} presents the experimental results (\eg, average accuracy) on relatively complex cross-domain incremental learning setting where datasets (BloodMNIST, PathMNIST, OrganaMNIST, and TissueMNIST) with varying modalities from different institutions are presented at each novel task. Results show an unmatched gap between regularization-based methods (\eg, Lwf and EWC) and our proposed distillation method. CL3DMC outperforms ER method by around 16\% on both task-aware and task-agnostic settings. Similarly, CL3DMC performs around 3\% better than the second best method, iCARL. 

\vspace{-3.0ex}    

\section{Conclusion}
\label{sec:conclusion}
In this paper, we propose a novel distillation strategy, L3DMC on mixed-curvature space to preserve the complex geometric structure  of medical data while training a DNN model on a sequence of tasks. L3DMC aims to optimize the lifelong learning model by minimizing the distance between new embedding and old subspace generated using current and old models respectively on higher dimensional RKHS. Extensive experiments show that L3DMC outperforms state-of-the-art L3 methods on standard medical image datasets for disease classification. In future, we would like to explore the effectiveness of our proposed distillation strategy on long-task and memory-free L3 setting.

\section*{Acknowledgments}
P.M. and K.R. gratefully acknowledge co-funding of the project by the CSIRO's Machine Learning and Artificial Intelligence Future Science Platform (MLAI FSP). K.R. also acknowledges funding from the CSIRO's ResearchPlus Postgraduate Scholarship. M.H. gratefully acknowledges the support from the Australian Research Council (ARC), project DP230101176.

\bibliographystyle{splncs04}
\bibliography{refs}

\title{L3DMC: Lifelong Learning using Distillation via Mixed-Curvature Space (Supplementary Material)}
\titlerunning{L3DMC}

\author{Kaushik Roy %
\inst{1, 2}, Peyman Moghadam %
\inst{2}, Mehrtash Harandi %
\inst{1}}
\authorrunning{K. Roy et al.}
\institute{Department of Electrical and Computer Systems Engineering, Faculty of Engineering, Monash University, Melbourne, Australia. \and  Data61, CSIRO, Brisbane, QLD, Australia.\\
\email{\{Kaushik.Roy, Mehrtash.Harandi\}@monash.edu \\
\{Kaushik.Roy, Peyman.Moghadam\}@csiro.au}}

\maketitle

\noindent
This document presents additional details that were omitted from
the main paper, due to the space constraints.

\paragraph{\textbf{Poincar\`e Ball.}} Exponential mapping function, $\text{exp}_{\vec{x}}^c$ offers a way to embed Euclidean feature point to Hyperbolic space (\eg, Poincar\`e Ball) and is defined as follows:
\begin{align}
\label{eq:exp_map}
    \exp_{\vec{x}}^c(\vec{v}) = \vec{x}~\oplus_c~\Big( \mathrm{tanh}\big(\sqrt{c}\frac{\lambda_{\vec{x}}^c\|\vec{v}\|}{2}\big)\frac{\vec{v}}{\sqrt{c}\|\vec{v}\|} \Big)
\end{align}

\paragraph{\textbf{Proof of the Proposed Distillation Loss.}}
In the main paper, we state that our distillation approach aims to minimizes the RKHS kernel interpolation error for preserving the geometric structure from prior model into current model. Assuming 
$\vec{Z}_{t-1}^{\mathfrak{e}} \in \mathbb{R}^{d \times m} = \{\vec{z}^{\mathfrak{e}}_{t-1,1}, \vec{z}^{\mathfrak{e}}_{t-1,2}, \cdots, \vec{z}^{\mathfrak{e}}_{t-1,m}\}$
be the output of the Euclidean projection module for $m$ samples using the old model and 
 $\vec{z}^{\mathfrak{e}}_{t} \in \mathbb{R}^{d \times 1}$ be a sample at time $t$ from the Euclidean projection head using current model, we proposed to minimize the following distance
\begin{equation}
\label{eqn:dist_loss}
  \delta^{\mathfrak{e}}(\vec{z}^{\mathfrak{e}}_{t}, \vec{Z}_{t-1}^{\mathfrak{e}}) = \hspace{-1mm}
    \min_{\alpha \in \mathbb{R}^m} \Big\| \phi(\vec{z}^{\mathfrak{e}}_{t}) - \sum_{i=1}^m \alpha_i \phi(\vec{z}^{\mathfrak{e}}_{t-1,i})\Big \|^2 = k(\vec{z}^{\mathfrak{e}}_{t},\vec{z}^{\mathfrak{e}}_{t}) - k_{\vec{z}\vec{Z}}^\top{K^{-1}_{\vec{Z}\vec{Z}}}k_{\vec{z}\vec{Z}}
\end{equation}
Bellow we provide the proof of Eq.~\eqref{eqn:dist_loss}. \\
\noindent
Since, $\phi$ in Eq.~\eqref{eqn:dist_loss} refers to the implicit mapping function to the RKHS defined by the Gaussian RBF kernel, \ie $k^{\mathfrak{e}}$, the distance can be represented as follows
\begin{align}
    \label{eqn:kernel_dist_subspace_}
    \Big\| \phi(\vec{z}^{\mathfrak{e}}_{t}) - \sum_{i=1}^m \alpha_i \phi(\vec{z}^{\mathfrak{e}}_{t-1,i})\Big \|^2\ = k(\vec{z}^{\mathfrak{e}}_{t},\vec{z}^{\mathfrak{e}}_{t}) - 2\cdot \alpha^\top k_{\vec{z}\vec{Z}} + \alpha^\top K_{\vec{Z}\vec{Z}}\alpha\;.
\end{align}
Performing derivative w.r.t. $\alpha$ and setting to zero results into $\alpha = K_{\vec{Z}\vec{Z}}^{-1}k_{\vec{z}\vec{Z}}$. Therefore, we can rewrite Eq.~\eqref{eqn:kernel_dist_subspace_} as follows
\begin{align}
    \label{eqn:kernel_dist_subspace_1}
    & \min_{\alpha \in \mathbb{R}^m} \Big\| \phi(\vec{z}^{\mathfrak{e}}_{t}) - \sum_{i=1}^m \alpha_i \phi(\vec{z}^{\mathfrak{e}}_{t-1,i})\Big \|^2\ \notag \\ 
    & = k(\vec{z}^{\mathfrak{e}}_{t},\vec{z}^{\mathfrak{e}}_{t}) - 2\cdot \{K_{\vec{Z}\vec{Z}}^{-1}k_{\vec{z}\vec{Z}}\}^\top K_{\vec{Z}\vec{z}} + \{K_{\vec{Z}\vec{Z}}^{-1}k_{\vec{z}\vec{Z}}\}^\top K_{\vec{Z}\vec{Z}}\{K_{\vec{Z}\vec{Z}}^{-1}k_{\vec{z}\vec{Z}}\} \notag \\
    & = k(\vec{z}^{\mathfrak{e}}_{t},\vec{z}^{\mathfrak{e}}_{t}) - 2\cdot k_{\vec{z}\vec{Z}}^\top \{K_{\vec{Z}\vec{Z}}^\top\}^{-1} k_{\vec{z}\vec{Z}} + k_{\vec{z}\vec{Z}}^\top \{K_{\vec{Z}\vec{Z}}^\top\}^{-1} K_{\vec{Z}\vec{Z}}K_{\vec{Z}\vec{Z}}^{-1}k_{\vec{z}\vec{Z}}
    \;.
\end{align}
Since $\{\vec{K_{\vec{Z}\vec{Z}}}^\top\}^{-1}=\{\vec{K_{\vec{Z}\vec{Z}}}^{-1}\}^\top$ and $\vec{K_{\vec{Z}\vec{Z}}}^\top = \vec{K_{\vec{Z}\vec{Z}}}$, $\{\vec{K_{\vec{Z}\vec{Z}}}^\top\}^{-1} \vec{K_{\vec{Z}\vec{Z}}} = \mathbb{I}$. Hence, Eq.~\eqref{eqn:kernel_dist_subspace_1} can be represented as follows

\begin{align}
    \label{eqn:kernel_dist_subspace_2}
    \min_{\alpha \in \mathbb{R}^m} \Big\| \phi(\vec{z}^{\mathfrak{e}}_{t}) - \sum_{i=1}^m \alpha_i \phi(\vec{z}^{\mathfrak{e}}_{t-1,i})\Big \|^2\ & = k(\vec{z}^{\mathfrak{e}}_{t},\vec{z}^{\mathfrak{e}}_{t}) - 2\cdot k_{\vec{z}\vec{Z}}^\top K_{\vec{Z}\vec{Z}}^{-1} k_{\vec{z}\vec{Z}} + k_{\vec{z}\vec{Z}}^\top K_{\vec{Z}\vec{Z}}^{-1}k_{\vec{z}\vec{Z}} \notag \\
    & = k(\vec{z}^{\mathfrak{e}}_{t},\vec{z}^{\mathfrak{e}}_{t}) - k_{\vec{z}\vec{Z}}^\top K_{\vec{Z}\vec{Z}}^{-1} k_{\vec{z}\vec{Z}}
    \;.
\end{align}

\paragraph{\textbf{Gradient of the Proposed Distillation Loss.}} In this section, we provide a closed form solution to computed gradient of our proposed knowledge distillation loss via mixed-curvature space.\\
\noindent
Gradient computation of our proposed distillation loss (Eq.~\eqref{eqn:dist_loss}), $\delta^{\mathfrak{e}}(\vec{z}^{\mathfrak{e}}_{t}, \vec{Z}_{t-1}^{\mathfrak{e}})$ with respect to $\vec{z}^{\mathfrak{e}}_{t}$ requires to compute $\frac{\partial \delta^{\mathfrak{e}}(\vec{z}^{\mathfrak{e}}_{t}, \vec{Z}_{t-1}^{\mathfrak{e}})}{\partial k_{\vec{z}\vec{Z}}}$. Using chain rule, we have
\begin{align}
\label{eqn:grad_e}
       \frac{\partial \delta^{\mathfrak{e}}(\vec{z}^{\mathfrak{e}}_{t}, \vec{Z}_{t-1}^{\mathfrak{e}})}{\partial \vec{z}^{\mathfrak{e}}_{t}} = 
       \frac{\partial \delta^{\mathfrak{e}}(\vec{z}^{\mathfrak{e}}_{t}, \vec{Z}_{t-1}^{\mathfrak{e}})}{\partial k_{\vec{z}\vec{Z}}}
       \frac{\partial k_{\vec{z}\vec{Z}}}{\partial \vec{z}^{\mathfrak{e}}_{t}}\;.
\end{align}
Now, 
\begin{align}
\label{eqn:grad_e_1}
       \frac{\partial \delta^{\mathfrak{e}}(\vec{z}^{\mathfrak{e}}_{t}, \vec{Z}_{t-1}^{\mathfrak{e}})}{\partial k_{\vec{z}\vec{Z}}} = 
       -2{K^{-1}_{\vec{Z}\vec{Z}}}k_{\vec{z}\vec{Z}}\;.
\end{align}
Considering an RBF kernel $k^{\mathfrak{e}}(\vec{z_{t}}, \vec{z_{t-1}}) \coloneqq \mathrm{exp}\big( -\lambda \| \vec{z_{t}} - \vec{z_{t-1}} \|^2 \big); \lambda > 0.$, $\big( \vec{z}^{\mathfrak{e}}_{t} - \vec{Z}_{t-1}^{\mathfrak{e}} \big)$ is as follows
\begin{align}
\label{eqn:grad_e_2}
    \frac{\partial k_{\vec{z}\vec{Z}}}{\partial \vec{z}^{\mathfrak{e}}_{t}} = -2\lambda k_{\vec{z}\vec{Z}} \big( \vec{z}^{\mathfrak{e}}_{t} - \vec{Z}_{t-1}^{\mathfrak{e}} \big)\;.
\end{align}
Combining all together, given an RBF function as kernel, the derivative of $\delta^{\mathfrak{e}}(\vec{z}^{\mathfrak{e}}_{t}, \vec{Z}_{t-1}^{\mathfrak{e}})$ with respect to $z_t^{\mathfrak{e}}$ for zero-curvature Euclidean space is as follows
\begin{align}
   \frac{\partial \delta^{\mathfrak{e}}(\vec{z}^{\mathfrak{e}}_{t}, \vec{Z}_{t-1}^{\mathfrak{e}})}{\partial \vec{z}^{\mathfrak{e}}_{t}} = 4\lambda d_{zZ}{K^{-1}_{\vec{Z}\vec{Z}}} k_{\vec{z}\vec{Z}}^2;~d_{zZ}=\big( \vec{z}^{\mathfrak{e}}_{t} - \vec{Z}_{t-1}^{\mathfrak{e}} \big)
\end{align}
For $d$-dimensional projected embedding space and $m$ embedding representation from old model, we have $d_{zZ} \in \mathbb{R}^{d \times m}$, $K^{-1}_{\vec{Z}\vec{Z}} \in \mathbb{R}^{m \times m}$ and $K_{\vec{z}\vec{Z}} \in \mathbb{R}^{m \times 1}$.

\paragraph{\textbf{Experimental Results.}}

\begin{figure*}[h!]
\vspace{-2.25em}
\centering
\begin{subfigure}{.33\textwidth}
  \centering
  \includegraphics[width=\linewidth, height=3cm]{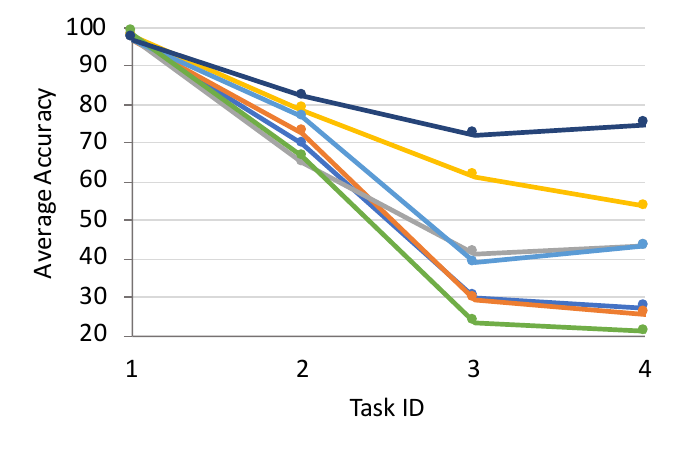}
  \caption{BloodMNIST}
  \label{fig:server}
\end{subfigure}%
\begin{subfigure}{.30\textwidth}
  \centering
  \includegraphics[width=\linewidth, height=3cm]{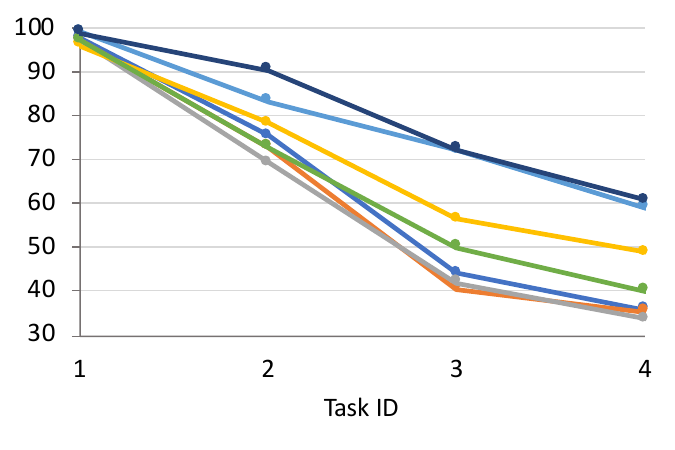}
  \caption{PathMNIST}
  \label{fig:client}
\end{subfigure}
\begin{subfigure}{.36\textwidth}
  \centering
  \includegraphics[width=\linewidth, height=3cm]{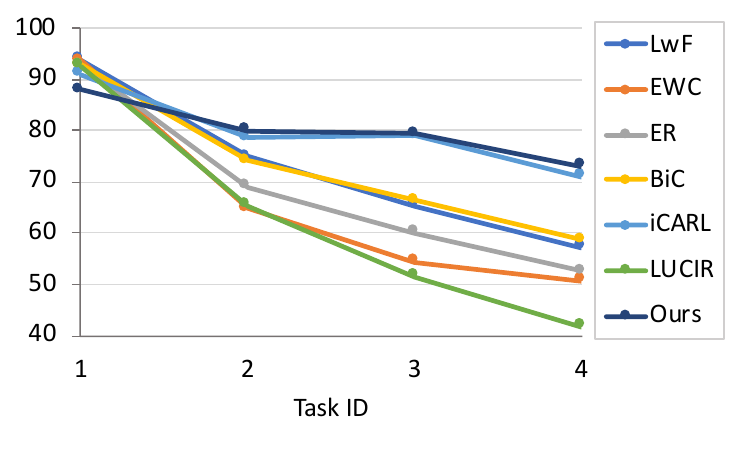}
  \caption{OrganaMNIST}
  \label{fig:client}
\end{subfigure}
\caption{Average accuracy in different tasks in 4-tasks class-incremental setting on BloodMNIST, PathMNIST and OrganaMNIST datasets with 200 exemplars in memory.}
\label{fig:inc_accuracy}
\vspace{-1.75em}
\end{figure*}

Figure~\ref{fig:inc_accuracy} shows the evolution of average accuracy of different methods on (a) BloodMNIST, (b) PathMNIST and (c) OrganaMNIST datasets with 200 memory samples. We observe that our proposed method consistently outperforms other baseline methods in consecitive tasks throughout the learning experience.

\end{document}